\documentclass[]{elsarticle}
\usepackage[T1]{fontenc}
\usepackage[top=1.5in,bottom=1.5in,left=1.in,right=1.in]{geometry}
\usepackage{hyperref}
\hypersetup{colorlinks,linkcolor={blue},citecolor={blue},urlcolor={blue}}
\usepackage[nowatermark]{fixmetodonotes}
\usepackage{amsfonts, amsmath, amssymb}
\usepackage{bbold}
\usepackage{booktabs}
\usepackage{array}
\usepackage{fourier}
\usepackage{longtable}
\usepackage{makecell}
\usepackage{verbatim}
\usepackage{siunitx}
\usepackage{microtype}
\usepackage[etex=true,export]{adjustbox}

\begin{document}

\begin{frontmatter}

\title{Self-supervised multimodal neuroimaging yields predictive representations for a spectrum of Alzheimer's phenotypes}

\author[trends]{Alex Fedorov}\corref{mycorrespondingauthor}
\cortext[mycorrespondingauthor]{Corresponding author}
\ead{afedorov@gatech.edu}

\author[trends]{Eloy Geenjaar}
\author[trends]{Lei Wu}
\author[borealisai]{Tristan Sylvain}
\author[trends]{Thomas P. DeRamus}
\author[mila]{Margaux Luck}
\author[trends]{Maria Misiura}
\author[mila,apple]{R~Devon Hjelm}
\author[trends]{Sergey M. Plis\fnref{fn1}}
\author[trends]{Vince D. Calhoun\fnref{fn1}}

\address[trends]{Tri-Institutional Center for Translational Research in Neuroimaging and Data Science, Georgia State, Georgia Tech, Emory, Atlanta, GA, USA}
\address[mila]{Mila - Quebec AI Institute, Montréal, QC, Canada}
\address[apple]{Apple Machine Learning Research, Seattle, WA, USA}
\address[umontreal]{Université de Montréal, Montréal, QC, Canada}
\address[borealisai]{Borealis AI, Montréal, QC, Canada}
\fntext[f1]{These two authors contributed equally}

\begin{abstract}
    Recent neuroimaging studies that focus on predicting brain disorders via modern machine learning approaches commonly include a single modality and rely on supervised over-parameterized models.
    However, a single modality provides only a limited view of the highly complex brain.
    Critically, supervised models in clinical settings lack accurate diagnostic labels for training.
    Coarse labels do not capture the long-tailed spectrum of brain disorder phenotypes, which leads to a loss of generalizability of the model that makes them less useful in diagnostic settings.
  This work presents a novel multi-scale coordinated framework for learning multiple representations from multimodal neuroimaging data.
  We propose a general taxonomy of informative inductive biases to capture unique and joint information in multimodal self-supervised fusion.
  The taxonomy forms a family of decoder-free models with reduced computational complexity and a propensity to capture multi-scale relationships between local and global representations of the multimodal inputs.
  Furthermore, we provide a methodology for analyzing learned inductive biases in deep learning models via downstream classification tasks, representation alignment, and saliency measures.
  We conduct a comprehensive evaluation of the taxonomy using functional and structural magnetic resonance imaging (MRI) data across a spectrum of Alzheimer's disease phenotypes and show that self-supervised models reveal disorder-relevant brain regions---including precuneus, hippocampus, and cingulate cortex---and multimodal links without access to the labels during pre-training.
  The proposed multimodal self-supervised learning yields representations with improved classification performance for both modalities.
  The representations also outperform unimodal supervised counterparts on fMRI data and advance the state-of-the-art in asymmetric coordinated fusion by exceeding the supervised model performance on one of the modalities.
  The concomitant rich and flexible unsupervised deep learning framework captures complex multimodal relationships and provides predictive performance that meets or exceeds that of a more narrow supervised classification analysis.
  We present elaborate quantitative evidence of how this framework can significantly advance our search for missing links in complex brain disorders.

\end{abstract}


\end{frontmatter}


\section{Introduction}
The brain is a vastly complex organ whose proper function relies on the simultaneous operation of multitudes of distinct biological processes. As a result, individual imaging techniques often capture only a single facet of the information necessary to understand a dysfunction or perform a diagnosis. As an illustration, structural MRI (sMRI) captures static but relatively precise anatomy, while fMRI measures the dynamics of hemodynamic response but with substantial noise.
Brain imaging analyses with a single modality have been shown to potentially lead to misleading conclusions~\cite{calhoun2016multimodal, plis2011effective}, which is unsurprising given fundamental differences in measured
information of the modalities as all of the modalities are flawed on their own in some way.

To address the limitations of unimodal analyses, it is natural to turn to multimodal data to leverage a wealth of complementary information, which is key to enhancing our knowledge of the brain and developing robust biomarkers.
Unfortunately, multimodal modeling is often a challenge, as finding points of convergence between different multimodal views of the brain is a  nontrivial problem. We propose self-supervised approaches to improve our ability to model joint information between modalities, thereby allowing us to achieve the following three goals.
Our primary goal in addressing multimodal modeling is to understand how to represent multimodal neuroimaging data by exploiting unique and joint information in two modalities.
As the second goal, we want to understand the links between different modalities (e.g., T1-weighted structure measurements and resting functional MRI data).
The final goal is to understand how to exploit
co-learning~\cite{baltruvsaitis2018multimodal}, especially in case one modality is particularly hard to learn.
By achieving these three goals in this work, we address three out of the five multimodal challenges~\cite{baltruvsaitis2018multimodal}: representation, alignment, and co-learning leaving only generative translation and fused prediction for future work.
Moreover, we present a general framework for self-supervised multimodal neuroimaging.
The proposed approach can capitalize on the available joint information to show competitive performance relative to supervised methods.
Our approach opens the door to additional data discovery. It enables characterizing subject heterogeneity in the context of imperfect or missing diagnostic labels and, finally, can facilitate visualization of complex relationships.

\subsection{Related work}

In neuroimaging, linear independent component analysis (ICA)~\cite{comon1994independent}, and canonical correlation analysis (CCA)~\cite{hotelling1992relations} are commonly used for latent representation learning and inter-modal link investigation.
Joint ICA (jICA)~\cite{moosmann2008joint} performs ICA on concatenated representation for each modality.
jICA has been extended with multiset canonical correlation plus ICA (mCCA+ICA)~\cite{sui2011discriminating} and spatial CCA (sCCA)~\cite{sui2010cca} which mitigate limitations of jICA or CCA applied separately.
These jICA or jICA-adjacent methods all estimate a joint representation.
Another approach, parallel ICA (paraICA)~\cite{liu2009combining}, simultaneously learns independent components for fMRI and SNP data that maximize the correlation between specific multimodal pairs of columns, in the mixing matrix of different modalities.
A recent improvement over paraICA, aNy-way ICA~\cite{duan2020any}, can scale to any number of modalities and requires fewer assumptions.

Most of the available multimodal imaging analysis approaches, including those mentioned above, rely on linear decompositions of the data.
However, recent work suggests the presence of nonlinearities in neuroimaging data that can be exploited by deep learning (DL)~\cite{abrol2021deep}.
Likewise, correspondence between modalities is unlikely to be linear~\cite{calhoun2016multimodal}.
These findings motivate the need for deep nonlinear models.
Within neuroimaging, supervised DL models have mostly proven to be successful due to their ease of use. These models show
mostly unparalleled performance mainly in the abundance of training data: a data sample paired with a corresponding label.
However, supervised models are prone to the shortcut learning phenomenon~\cite{geirhos2020shortcut}; when a model hones in on trivial patterns in the training set that are sufficient to classify the available data but are not generalizable to data unseen at training.
Next, it has been shown that supervised models can memorize noisy labels~\cite{arpit2017closer} which are commonplace in healthcare~\cite{pechenizkiy2006class,rokham2020addressing}.
Furthermore, supervised methods have also been shown to be data-inefficient~\cite{CPCv2} while labels in medical studies are costly and scarce.
Finally, in many cases, diagnostic labels are based on self-reports and interviews and thus may not accurately reflect the underlying biology~\cite{rokham2020addressing}. Many of these problems can be addressed with unsupervised learning and, more recently, self-supervised learning (SSL)~\cite{dosovitskiy2014discriminative}.
In SSL, a model trains on a proxy task that does not require externally provided labels.
SSL has been shown to improve robustness~\cite{NEURIPS2019_a2b15837}, data-efficiency~\cite{CPCv2}, and can outperform supervised approaches on image recognition tasks~\cite{SWAV}.

In the early days of the current wave of unsupervised deep learning, common approaches were based on deep belief
networks (DBNs)~\cite{srivastava2012learning, plis2014deep}, and deep Boltzmann machines (DBMs)~\cite{srivastava2012multimodal, hjelm2014restricted, SUK2014569}.
However, DBNs and DBMs are difficult to train.
Later, deep canonical correlation analysis (DCCA)~\cite{dcca} was introduced for multiview unsupervised learning.
DCCA~\cite{dcca}, and its successor, deep canonically correlated autoencoder (DCCAE)~\cite{DCCAE} are
trained in a two-stage procedure. In the first stage, a neural network trains unimodally via layer-wise pretraining or using an autoencoder. In the second stage, CCA is used to capture joint information between modalities.
Due to the need for a decoder, the autoencoding approaches demand high computational and memory requirements for full brain data as most brain segmentation models are still working on 3D patches~\cite{fedorov2017end,fedorov2017almost,henschel2020fastsurfer}.

Among many self-supervised learning approaches, we are specifically interested in methods that use maximization of mutual information, such as Deep Infomax (DIM)~\cite{DIM} and contrastive predictive coding (CPC)~\cite{CPCv1}.
These methods can naturally be extended to modeling multimodal data~\cite{fedorov2021self} compared to other self-supervised pre-text tasks~\cite{misra2020self} (e.g. relative position~\cite{doersch2015unsupervised}, rotation~\cite{gidaris2018unsupervised}, colorization~\cite{zhang2016colorful}).
The maximization of mutual information in these methods allows a predictive relationship between representations at different levels as a learning signal for training.
Specifically, the learning signal in DIM~\cite{DIM} is the relationship between the intermediate representation of a convolutional neural network (CNN) and the whole representation of the input. In (CPC)~\cite{CPCv1}, for example, this is done between the context and a future intermediate state.
Both DIM and CPC have been successfully extended and applied unimodally for the prediction of Alzheimer's disease from sMRI~\cite{fedorov2019prediction}, transfer learning with fMRI~\cite{mahmood2019transfer, mahmood2020whole}, and brain tumor, pancreas tumor segmentation, and diabetic retinopathy detection~\cite{NEURIPS2020_d2dc6368}.
In addition, these models do not reconstruct as part of their learning objective, unlike autoencoders.
The reconstruction-free model saves a lot of compute and memory, especially for volumetric medical imaging applications.

In multiview and multimodal settings, self-supervised learning has enabled state-of-the-art results on various computer vision problems via maximization of mutual information between different views of the same image~\cite{amdim,cmc,simclr}, and multimodal data (e.g., visual, audio, text)~\cite{miech2020end, alayrac2020self, Radford2021LearningTV,CMIM}.
These SSL approaches capture the joint information between two corresponding distorted or augmented images, image-text, video-audio, or video-text pairs.
In the multiview case~\cite{amdim,cmc,simclr}, the models learn
transformation-invariant representations by capturing the joint
information while discarding information unique to a transformation.
In the case of multimodal data, the models learn modality-invariant~\cite{demian} representations known as retrieval models.
The same ideas have been extended to the multidomain scenario to learn domain-invariant~\cite{feng2019self} representations.
However, in the case of neuroimaging, when one modality can capture
the anatomy, and the other can capture brain dynamics, the joint information alone
will not be sufficient due to the information content that each of the modalities is measuring.
We hypothesize that we additionally need to capture unique modality-specific information.

Most of the described multiview, multidomain, and multimodal
work can be viewed as a coordinated representation
learning~\cite{baltruvsaitis2018multimodal}.
In terms of coordinated representation learning, we learn separate representations for each view, domain, or modality.
But the representations are coordinated through an objective function by optimizing a similarity measure with possible additional constraints (e.g., orthogonality in CCA).
In this case, the objective function mainly captures joint information between the \emph{global} latent representation of modalities that summarize the whole input.
However, such framework only considers a \emph{global}-\emph{global} relationship between modalities.
To resolve this limitation, we can consider intermediate representations, namely as \emph{local} representation that captures local information about the input.
That would allows us to capture \emph{local-to-local}, \emph{glocal-to-global},
\emph{local-to-global} multimodal relationships.
Previously, augmented multiscale DIM (AMDIM)~\cite{amdim}, cross-modal DIM (CM-DIM)~\cite{sylvain2019locality,sylvain2020zeroshot}, and spatio-temporal DIM (ST-DIM)~\cite{anand2019unsupervised} used \emph{local} intermediate representation of convolutional layers to capture multi-scale relationships between multiple views, modalities or time frames.
Thus, we hypothesize that multiscale relationships between modalities can also be used to learn representations from multimodal neuroimaging data.
We extend the coordinated representation learning framework to a multiscale coordinated representation framework for multimodal data to verify this.

\subsection{Contributions}

First, we propose a multiscale coordinated framework as a family of models.
The family of the models is inspired by many published SSL approaches based on the maximization of mutual information that we combine in a complete taxonomy.
The family of methods within this taxonomy covers multiple inductive biases that can capture joint inter-modal and unique intra-model information.
In addition, it covers multiscale multimodal relationships in the input data.

Secondly, we provide a methodology to exhaustively evaluate learned representation.
We thoroughly investigate the models on a multimodal dataset
OASIS-3~\cite{OASIS3} by evaluating performance on two classification
tasks, measuring the amount of joint information in representations
between modalities, and interpreting the representation in the brain voxel-space.

Our results provide strong evidence that self-supervised models yield useful predictive representations for classifying a spectrum of Alzheimer's phenotypes.
We show that self-supervised models are able to uncover regions of interest, such as the hippocampus~\cite{yang2022human}, thalamus~\cite{elvsaashagen2021genetic}, parietal lobule~\cite{greene2010subregions}, occipital gurys~\cite{yang2019study} on T1, and hippocampus~\cite{10.3389/fnagi.2018.00037,zhang2021regional}, middle temporal gyrus~\cite{hu2022brain}, subthalamus hypothalamus~\cite{chan2021induction} and superior medial frontal gyrus~\cite{cheung2021diagnostic} on fALFF.
Furthermore, self-supervised models can uncover multimodal links between anatomy and brain dynamics that are also supported by previous literature, such as Thalamus (on fALFF) - Precuneus (on T1)~\cite{cunningham2017structural}, precuneus (T1) and hippocampus (fALFF)~\cite{kim2013hippocampus, ryu2010measurement}, and precuneus (T1) and middle cingulate cortex (fALFF)~\cite{rami2012distinct, bailly2015precuneus}.

\section{Materials and Methods}
In this section, we first describe the foundation of our framework. Then we describe the evaluation of learned representation.

\subsection{Overview of coordinated representation learning}
To help the reader understand the foundation of our framework, we start with the idea of coordinated representation learning~\cite{baltruvsaitis2018multimodal}.

Let $x = (x^1, ..., x^M)$ be an arbitrary sample of $M$ related input
modalities collected from the same subject.
Taken together, a set of samples comprise a multimodal dataset $\mathcal{D} = \{(x_i^1, ..., x_i^M)\}_{i=1}^N$.
In our case, the modalities of interest are sMRI and rs-fMRI represented
as T1 and fALFF volumes, respectively. Thus, $M = 2$ and $x^m \in
R^{d_1\times d_2\times d_3}$, a $d_1\times d_2\times d_3$ tensor.
While inter-modal relationships can be as complex as a concurrent acquisition of
neuroimaging modalities, we pair sMRI and rs-fMRI simply based on the
temporal proximity between sessions of the same subject from the OASIS-3~\cite{OASIS3} dataset.

For each modality $m$, coordinated representation learning introduces an independent encoder $E^m$ that maps the input, $x^m$, to a lower $d$-dimensional representation vector $z^m = E^m(x^m)$.
Thus, each modality has a separate encoder and a corresponding representation.

In this work, we parameterize encoder $E^m$ for each modality $m$ with a volumetric deep convolutional neural network.
To learn each encoder's parameters, we optimize an objective function $\mathcal{L} = \mathcal{L}(z^1, ..., z^M)$ that incorporates the representation of each modality and coordinates them.
The objective encourages each of the encoders to learn an encoding for their respective modality that is informed by the other modalities. This cross-modal influence is learned during training and is captured in the parameters of each encoder.
Hence, a representation of an unseen subject's modality will capture cross-modal influences without a contingency on the availability of the other modalities.
One common choice with respect to cross-modal influences is to coordinate representations across
modalities $z^m$ by maximizing their similarity metric~\cite{frome2013devise} or correlation via CCA~\cite{dcca} between representation vectors.

To summarize, in coordinated representation learning,
modality-specific encoders learn to generate
representations in a cross-coordinated manner guided by an objective
function.

The main limitation of the coordinated representation framework is its exclusive focus on capturing joint information between modalities instead of also capturing information that is exclusive to each modality.
Thus, \emph{DCCA}~\cite{dcca} and \emph{DCCAE}~\cite{DCCAE} employ coordinated learning only as a secondary stage after pretraining the encoder.
The first stage in these methods focuses on learning modality-specific information via layer-wise pretraining of the encoder in \emph{DCCA}, or pretraining of the encoder with an autoencoder (\emph{AE}) in \emph{DCCAE}.
On the other hand, deep collaborative learning (DCL)~\cite{DCL} attempts to capture modality-specific information using a supervised objective for phenotypical information with respect to each modality, in addition to CCA.

An additional limitation that previous work has not considered is the use of intermediate representation in the encoder.
Intermediate representations have been integral to the success of U-Net architecture in biomedical image segmentation~\cite{ronneberger2015u}, brain segmentation tasks~\cite{henschel2020fastsurfer}, achieving state-of-the-art results with self-supervised learning on natural image benchmarks with DIM~\cite{DIM} and AMDIM~\cite{amdim}, and achieving near-supervised performance in Alzheimer's disease progression prediction, with self-supervised pretraining~\cite{fedorov2019prediction}.

In our work, we address these limitations by proposing a  multi-scale coordinated learning framework.

\subsubsection{Multi-scale coordinated learning}

To motivate multi-scale coordinated learning, we re-introduce intermediate representations and explain how they can benefit multimodal modeling.

Each encoder $E^m$ produces intermediate representations.
Specifically, if the encoder $E^m$ is a convolutional neural network (CNN) with $L$ layers, each subsequent layer $l=1,...,L$ in the CNN represents a larger part of the original input data.
Furthermore, each of these scales, which correspond to the depth of a layer, is an increasingly non-linear transformation of the input and produces a more abstract representation of that input relative to the previous scales.
The intermediate representations of layer $l$ are $S \times o$ convolutional features $\{c_l^m\}_{S_l}$, where $S$ is the number of locations in the convolutional features of layer $l$, and $o$ is the number of channels.
These features are also often referred to as activation maps within the network.
For example, if the input is a 3D cube, the arbitrary feature size $s$ within a layer $l$ of the CNN will be $s \times s \times s$.
Thus, each of the intermediate representations will have $S = s^3$ locations.
Each location in the intermediate representation has a receptive
field~\cite{araujo2019computing} that captures a certain subset of the input sample.
Each intermediate representation thus captures some of the input's \emph{local} information, while the latent representation ($z^m$) captures the input's \emph{global} information.

\begin{figure}[t]
    \centering
    \includegraphics[width=0.6\linewidth]{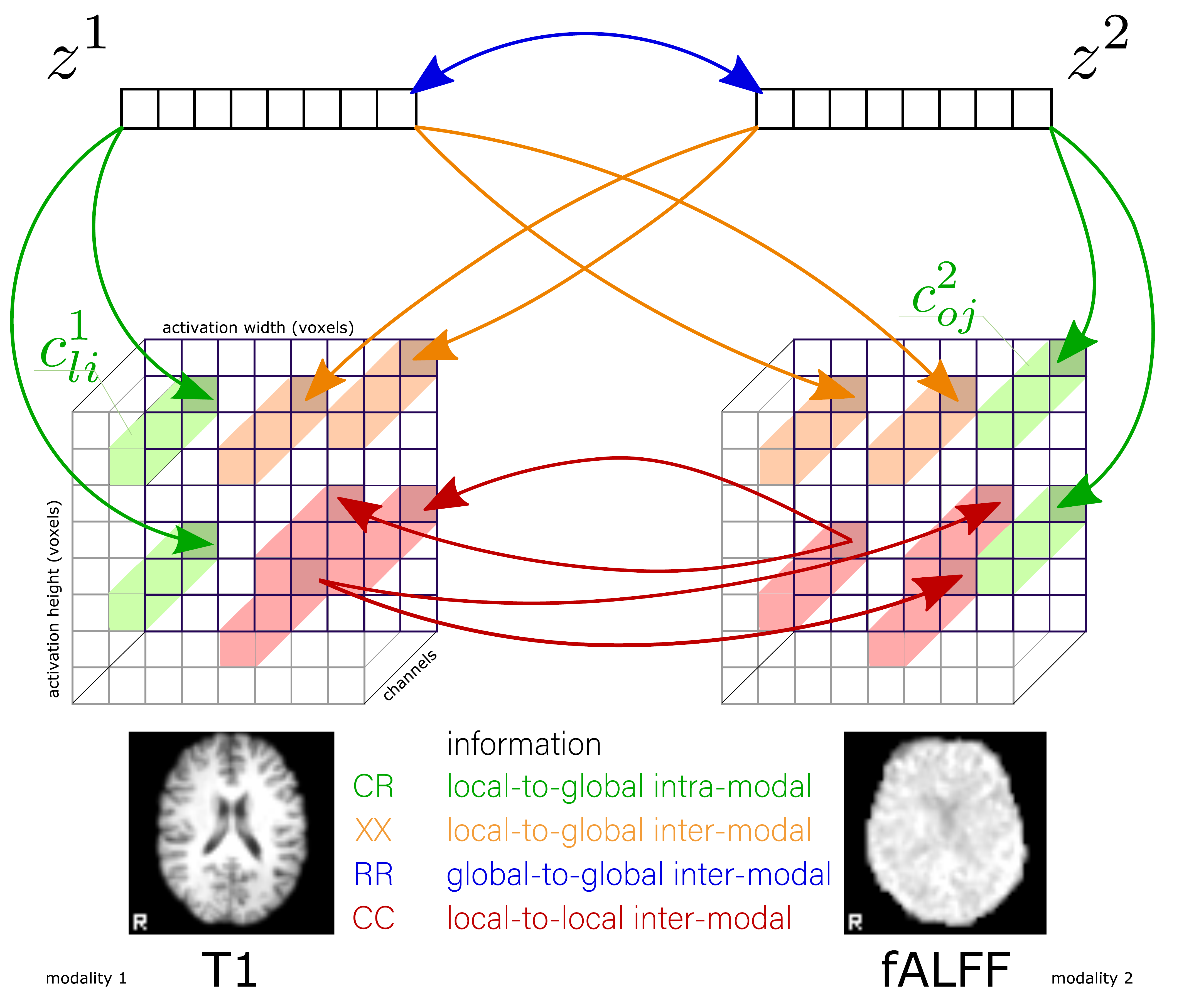}
    \caption{The concept behind the multi-scale coordinated learning based on four principle relationships: \emph{Convolution-to-Representation} (\emph{CR}), \emph{Cross Convolution-to-Representation} (\emph{XX}), \emph{Representation-to-Representation} (\emph{RR}), and \emph{Convolution-to-Convolution} (\emph{CC}). Each colored vecor in the convolution activation map $c^1_{li}$ and $c^2_{oj}$, corresponds to arbitrary locations $i$ and $j$ in features maps of layers $l$ and $o$ for modalities $1$ and $2$, respectively. We use ``\emph{local} representation'' to denote each location in the convolutional activation map: the $c$ vector spanning the channels. Latent representation vector $z$ is the d-dimensional \emph{global} representation. To avoid clutter we display only a slice of data but layer activations a volume per channel in our applications.}
    \label{fig:scheme}
\end{figure}

With two scales and two modalities, we can define multi-scale coordinated learning based on four objectives, which are schematically shown in Figure~\ref{fig:scheme}.
The \emph{Convolution-to-Representation} (CR) objective captures modality-specific information as \emph{local-to-global} intra-model interactions.
The \emph{Cross Convolution}\emph{-to-Representation} (XX) objective captures joint inter-modal \emph{local-to-global} interactions between the \emph{local} representations in one modality and the \emph{global} representation in another modality.
The \emph{Representation}\emph{-to-Representation} (RR) objective captures joint information between \emph{global} inter-modal representations as \emph{global-to-global} interactions.
The \emph{Convolution}\emph{-to-Convolution} (CC) objective captures joint information between \emph{local} inter-modal representations as \emph{local-to-local} interactions.

Thus, we can capture modality-specific information and multimodal relationships at multiple scales.
These extensions cover two of the previously mentioned limitations in the coordinated learning framework.
Our extensions also allow us to define a full taxonomy of models that can be constructed based on these four principal interactions and to show how these compare to or supersede related work.

To construct an effective objective $\mathcal{L}$ based on these multi-scale coordinated interactions, first, we will define an estimator of mutual information.
This estimator will be used to define each of the four objectives as a mutual information maximization problem that can be used to encourage the interactions between the corresponding representations.
Lastly, we explain how one can construct an objective
$\mathcal{L}$ for a multi-scale coordinated representation learning problem, based on a combination of the four basic objectives between \emph{global} and \emph{local} features, and, additionally, show how these compare to related work.

\subsubsection{Mutual Information Maximization}

To estimate mutual information between random variables $X$ and $Y$, we use a lower-bound, based on the noise-contrastive estimator (InfoNCE)~\cite{CPCv1}.
\begin{equation}
    \label{eq:infonce}
    I(X; Y) \ge I^{\text{InfoNCE}}(X; Y) = \frac{1}{N} \sum_{i=1}^N \log \frac{e^{f(x_i,y_i)}}{\frac{1}{N}\sum_{j=1}^N \mathbb{1}_{i \ne j}e^{f(x_i,y_j)}},
\end{equation}
where the samples $x_i \sim X, y_i \sim Y$ construct pairs: $(x_i, y_i)$ sampled from the joint
$P(X, Y)$ (\emph{positive} pair) and $(x_i, y_j)_{i \ne j}$ sampled
from the product of the marginals $P(X) \otimes P(Y)$ (\emph{negative}
pair).
The $x_i \in \mathbb{R}^d$ and $y_i \in \mathbb{R}^d$ represent $d$-dimensional representation vectors, and can be \emph{local} or \emph{global} representations.
The function $f: \mathbb{R}^d\rightarrow\mathbb{R}$ in equation \ref{eq:infonce} is a scoring function that maps its input vectors to a scalar value and is supposed to reflect the goodness of fit. This functions $f$ is also known as the \emph{critic function}~\cite{Tschannen2020On}.
The encoder is optimized to maximize the critic function for a positive pair and minimize it for a negative pair, such that $f(x_i,y_i) \gg f(x_i,y_j)_{i \ne j}$.
Our choice of the critic function is a scaled dot-product~\cite{amdim}, and is defined as:
\begin{equation}
    f(x,y) = \frac{x^{\intercal} y}{\sqrt{d}}
\end{equation}

\subsubsection{Taxonomy for multi-scale coordinated learning}

\begin{figure}[t]
  \centering
    \includegraphics[width=0.7\linewidth]{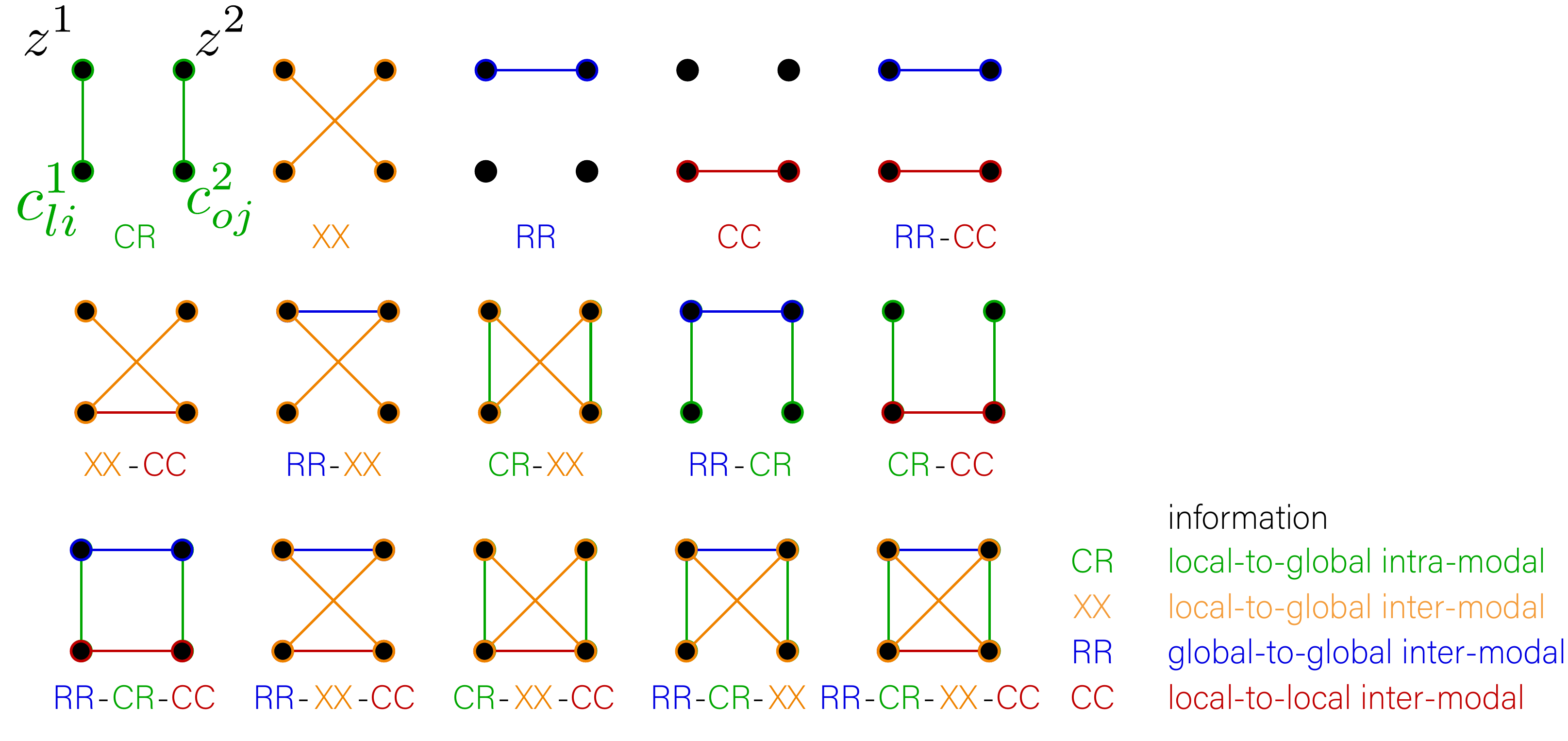}
    \caption{The complete taxonomy of interactions, based on the four principle interactions. The lower dots are the convolutional activations, whereas the upper dots are the global representations. The interactions are defined between 1st modality (left) and 2nd modality (right). The combinations represent names of the models that are based on these four interactions: \emph{CR}, \emph{XX}, \emph{RR} and \emph{CC}. The colors follow the colormap from Figure~\ref{fig:scheme}.}
    \label{fig:variants}
\end{figure}

Given the mutual information estimator, we can construct four basic objectives and then use those to construct a full taxonomy of interactions, which is shown in Figure~\ref{fig:variants}.
Each option within the taxonomy specifies a unique optimization objective $\mathcal{L(\mathcal{D})}$.
Notably, the first row of the figure shows the principal losses: \emph{CR}, \emph{XX}, \emph{RR}, and \emph{CC}, - that we defined before.
The remaining parts of the taxonomy are constructed by adding a composition of the principle losses together.
For example, the 5th combination\emph{RR-CC} is the sum of the two basic objectives \emph{RR} and \emph{CC}.

To discuss the options in the taxonomy, we first reintroduce some notations.
A \emph{local} representation $c^m_{ld}$ is any arbitrary location $d \in \{1,...,S\}$ in the convolutional feature $c_m^l$ from a convolutional layer $l$, with $S$ locations.
A location is represented as the $C$-dimensional vector, where $C$ is the number of channels of the convolutional activation map for layer $l$.
The choice of the layer $l$ is a hyperparameter and can be guided by the following intuition.
First, the chosen layer $l$ should not be too close to the last layer or be the last layer because it will capture similar information to the global $z$ representation.
In this case, the local and global content of the input could be very similar, with almost or exactly the same receptive field.
Although a single layer difference between the \emph{local} and \emph{global} representations can be used as a strategy for layer-wise pretraining, as in Greedy Deep InfoMax (GIM)~\cite{lowe2019putting}, and the least case is not meaningful.
Secondly, the chosen layer $l$ should not be too close to or be the first layer.
This will lead to a \emph{local} representation with a very small receptive field that only captures hyper-local information of the input.
The first layer will essentially capture the intensity of a voxel, which has not worked well, as previously shown in DIM~\cite{DIM}.

The \emph{global} representation is the encoder's $E^m$ latent representation $z^m = E^m(x^m)$ that summarizes the whole input.
The \emph{global} representation is a $d$-dimensional vector, where $d$ is also a hyperparameter.
With this \emph{global} representation, we also define a $d$-dimensional space, wherein we compute scores with the critic function $f$.
The the \emph{local} representation, however, is a $C$-dimensional vector. To overcome the difference in size, we add an additional \emph{local} projection head $\ell$.
This projection head takes the $C$-dimensional \emph{local} representation from layer $l$ in the encoder and projects it to the $d$-dimensional space so we can compute scores with $f$ in this $d$-dimensional space.
This projection is also parameterized by a neural network and separate for each modality.
In addition, we introduce a \emph{global} projection head $g$.
Both the \emph{local} and \emph{global} projection heads are shown to improve the training performance in DIM~\cite{DIM} and in SimCLR~\cite{simclr}, respectively.

The first objective (CR), in the top left corner of Figure~\ref{fig:variants}, trains two independent encoders, one for each modality, with a unimodal loss function that maximizes the mutual information between local $c^m_{ld}$ and global $z^m$ representations.
This objective directly implements the Deep InfoMax (DIM)~\cite{DIM} objective.
The idea behind this approach is to maximize the information between the lowest and the highest scales of the encoder.
In other words, the local representations are driven to be predictive of the global representation.
The objective for an arbitrary layer $l$ is defined as:
\begin{align}
\includegraphics[valign=c,width=0.05\linewidth]{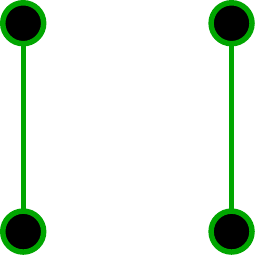} \qquad    {\cal
                                                     L}^{\text{CR}}(m)  =
                           I^{\text{InfoNCE}}(\ell^m(\{c^m_{ld}\}_{d=1,...,S});
                           g^m(\{z^m\})) = \frac{1}{N} \sum_{i=1}^N \log
                           \frac{e^{f(\ell^m(c^m_{i,ld}),g^m(z^m_i))}}{\frac{1}{N}\sum_{j=1}^N
                           \mathbb{1}_{[i \ne
                           j]}e^{f(\ell^m(c^m_{i,ld}),g^m(z^m_j))}},
\end{align}
, where we only define the following objective ${\cal L}^{\text{CR}}(m)$ for a modality $m$.
The objective has to be computed for each modality.

The CR objective can be extended to the multimodal case by measuring the
inter-modal mutual information between local $c^m_{ld}$ and global
$z^k$ representations of modality $m$ and $k \ne m$,
respectively.
We call this multimodal objective Cross Convolution-to-Representation (\emph{XX}), and it is shown second from the top left in Figure~\ref{fig:variants}.
This objective has previously been used in the context of augmented
multiscale DIM (AMDIM)~\cite{amdim}, cross-modal DIM
(CM-DIM)~\cite{sylvain2020zeroshot}, and spatio-temporal DIM
(ST-DIM)~\cite{anand2019unsupervised}. We define it as
\begin{align}
\includegraphics[valign=c,width=0.05\linewidth]{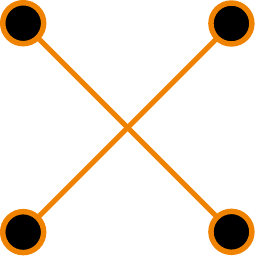} \qquad
{\cal L}^\text{XX}(m,k) = I^{\text{InfoNCE}}(\ell^m(\{c^m_{ld}\}_{d=1,...,S}); g^k(\{z^k\}))_{k \ne m},
\end{align}
, where the objective ${\cal L}^\text{XX}(m,k)$ is defined for a pair of modalities $m$ and $k$.
The objective has to be computed for all possible pairs of modalities.
In case of symmetric coordinated fusion, the symmetry has to be preserved for modalities $m$ and $k$ by computing both ${\cal L}^\text{XX}(m,k)$ and ${\cal L}^\text{XX}(k,m)$, whereas for asymmetric fusion, this is not the case.

The third elementary objective measures mutual information between the global representation of one modality $z^m$ and the global representation of another modality $z^k$, $k \ne m$.
This objective is called Representation-to-Representation (\emph{RR})
and is the third in the top row of Figure~\ref{fig:variants}.
This interaction has been used in many prior contrastive multiview
work~\cite{cmc,simclr,moco,SWAV} and \emph{DCCA})~\cite{dcca}.
The RR objective is defined as
\begin{align}
\includegraphics[valign=c,width=0.05\linewidth]{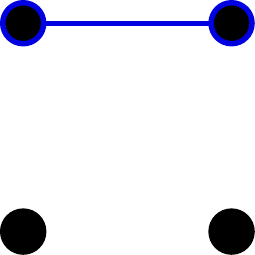} \qquad
  {\cal L}^{\text{RR}}(m,k) = I^{\text{InfoNCE}}(g^m(\{z^m\}); g^k(\{z^k\})),
\end{align}
, where we the objective ${\cal L}^{\text{RR}}(m,k)$ is defined for a pair of modalities $m$ and $k$.

The fourth elementary objective is similar to \emph{RR}, but only
maximizes the mutual information between the two inter-modal local
representations $c^m_{ld}$ and $c^k_{of}$, where $d$ and $f$ are arbitrary locations in layers $l,o$ within modality $m$ and $k \ne m$, respectively.
This objective is called Convolution-to-Convolution (\emph{CC}) and is shown as the fourth from the top left in Figure~\ref{fig:variants}.
The \emph{CC} objective has been used in AMDIM~\cite{amdim}, CM-DIM~\cite{sylvain2020zeroshot}, and ST-DIM~\cite{anand2019unsupervised}.
Due to a large number of possible pairs of locations between the activation maps in each encoder, we reduce the computational costs by sampling arbitrary locations, which was proposed in AMDIM~\cite{amdim}.
Thus, after sampling an arbitrary location from the convolutional activation map for one modality, we compute the objective in a similar way to XX, by treating sampled locations as the global representation.
\begin{align}
\includegraphics[valign=c,width=0.05\linewidth]{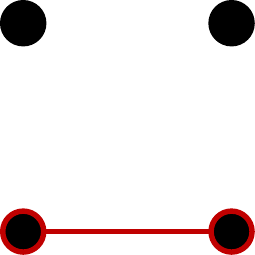} \qquad     {\cal L}^{\text{CC}}(m,k)= I^{\text{InfoNCE}}(\ell^m(\{c^m_{ld}\}_{d=1,...,S}); \ell^k(\{c_{l*}^k\}))_{k \ne m},
\end{align}
, where $* \sim {1,...,T}$ is a sampled location from $T$ locations.
The objective ${\cal L}^{\text{CC}}(m,k)$ is defined for a pair of modalities $m$ and $k$.

By combining these four primary objectives, we can construct more complicated objectives, as shown in Figure~\ref{fig:variants}.
For example, the \emph{XX-CC} objective for two modalities (as $m = \text{M1}$ and $k = \text{M2}$) can be written as
\begin{align}
\includegraphics[valign=c,width=0.05\linewidth]{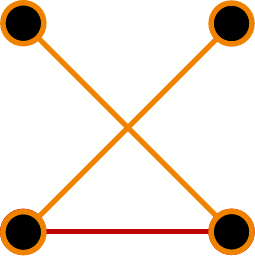} \qquad \mathcal{L}^{\text{XX-CC}}(D^{\text{M1}}, D^{\text{M2}}) = {\cal
L}^{XX}(\text{M1}, \text{M2}) + {\cal L}^{XX}(\text{M2},\text{M1}) + {\cal L}^{CC}(\text{M1},\text{M2}) + {\cal L}^{CC}(\text{M2},\text{M1})
\end{align}
The goal would be to find parameters $\Theta$ that maximize $\mathcal{L}^{\text{XX-CC}}$.
The objective is repeated with the modalities flipped to preserve the symmetry of \emph{XX} and \emph{CC}.
Removing the symmetry is intuitively similar to guiding the representations of one modality by the representations of another modality, which may be interesting for future work on asymmetric fusion.
The \emph{XX-CC} objective coordinates representations locally with the \emph{CC} objective on convolutional activation maps and coordinates representations across scales in the encoder with \emph{XX}. The local representations of one modality should be predictive of the global and local representations of the other modality.

\subsubsection{Baselines and other objectives}

\begin{figure}[t]
    \centering
    \includegraphics[width=0.5\linewidth]{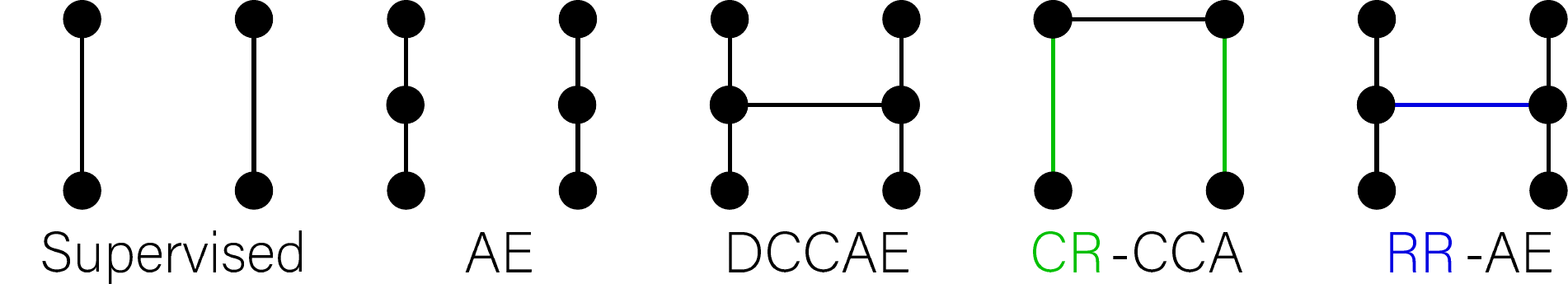}
    \caption{This figure shows schemes for the following models: \emph{Supervised}, autoencoder (\emph{AE}), deep canonical correlation autoencoder (\emph{DCCAE}), the \emph{CR} objective combined with \emph{CCA} (\emph{CR-CCA}), and the \emph{RR} objective combined with an autoencoder (\emph{RR-AE}).
    The Supervised and \emph{CR-CCA} objectives follow a similar structure as schemes represented in our taxonomy, see Figure~\ref{fig:variants}.
    The AE-based models (\emph{AE}, \emph{DCCAE}, \emph{RR-AE}) represented by 3-dot scheme where the middle dot is a representation, the lower dots are convolutional activation maps of the input, and the upper dots are reconstructions of the input.}
    \label{fig:other_variants}
\end{figure}

We compare our method to an autoencoder (\emph{AE}), a deep canonical correlation autoencoder (\emph{DCCAE})~\cite{DCCAE}, and a supervised model. Each type of model is a high-performing version of the three main categories of alternative approaches to our framework.
The \emph{AE} and supervised models are trained separately for each modality, while the \emph{DCCAE} is trained jointly on all modalities.
By \emph{Supervised}, we refer to a unimodal model that is trained to predict a target using cross-entropy loss.

In addition to defining a unified framework that covers multiple existing approaches, our taxonomy contains a novel unpublished approach that combines different combinations of the four objectives.
One novel approach combines the \emph{CR} objective with the objective of the \emph{DCCAE}, which we call \emph{CR-CCA}.
The \emph{CR} objective allows us to train using modality-specific information, and the \emph{CCA} objective aligns the representations between modalities. This leads to the following objective:
\begin{align}
  \includegraphics[valign=c,width=0.05\linewidth]{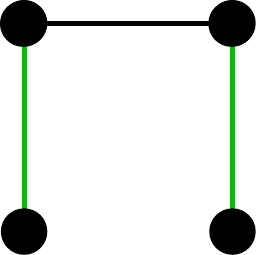}
  \qquad
  \mathcal{L}^{\text{CR-CCA}}(D^{\text{M1}}, D^{\text{M2}}) = {\cal
L}^{CCA}(\text{M1}, \text{M2}) + {\cal L}^{CCA}(\text{M2},\text{M1}) + {\cal L}^{CR}(\text{M1},\text{M2}) + {\cal L}^{CR}(\text{M2},\text{M1})
\end{align}

A second novel approach combines the \emph{AE} objective with our \emph{RR} objective to create the \emph{RR-AE} objective.
The \emph{AE} objective ensures the learning of modality-specific representations, and the \emph{RR} objective enforces the alignment of representations across modalities, similar to the \emph{CCA} objective in the \emph{DCCAE}.
The final objective of the \emph{RR-AE} is as follows.
\begin{align}
    \includegraphics[valign=c,width=0.05\linewidth]{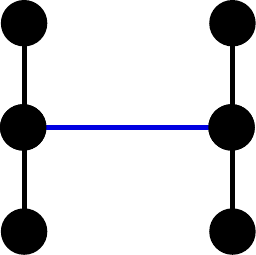}
  \qquad
  \mathcal{L}^{\text{RR-AE}}(D^{\text{M1}}, D^{\text{M2}}) = {\cal
L}^{RR}(\text{M1}, \text{M2}) + {\cal L}^{RR}(\text{M2},\text{M1}) - R^{M1} - R^{M2},
\end{align}
where $R^M$ is the mean squared reconstruction error for the AE with an additional decoder $D$, and modality $M$:
\begin{equation}
    R^M = \frac{1}{N}  \sum_{n=1}^N ||x^M_n - D^M(E^M(x^M_n))||^2
\end{equation}
The baseline schemes are shown in Figure~\ref{fig:other_variants}.

\subsection{Analysis of the model representation}
It is important to validate the representations that the model learns, which in the proposed framework is done in three steps. The first step evaluates the representations using classification tasks with logistic regression. This step uses the features that are extracted by the encoder from the input. The goal of this evaluation is to ensure the discriminative power of the pre-trained features. In the second step, we compute the similarity between representations to measure how much joint information has been captured by the model. The third and last step consists of two analyses to explore the relationship between the latent space and the brain space to assess voxel-wise group differences based on saliency gradients for each of the $d$ dimensions of the representation.

\subsubsection{Classification evaluation}

To evaluate the discriminative performance of the representations captured by the model, we train logistic regression on frozen representations from the last layer of the encoder.
Note that most self-supervised learning algorithms evaluate the discriminative power of representations with a linear evaluation protocol based on linear probes~\cite{linear_probes}.
We chose to use logistic regression, however, due to faster training times.

\subsubsection{Alignment analysis}
To evaluate the alignment between representations of different modalities, we use central kernel alignment (CKA)~\cite{CKA}.
CKA is shown to be effective~\cite{CKA} as a method to identify the correspondence between representations of networks with different initializations, compared to CCA-based similarity measures~\cite{SVCCA, PWCCA}.
CKA is considered to be a normalized version of the Hilbert-Schmidt Independence Criterion (HSIC)~\cite{gretton2005measuring}. The CKA measure for a pair of modalities $m$ and $k$ is defined as:
\begin{equation}
\text{CKA}(Z^m, Z^k) = \frac{\left\|Z^{k\mathrm{T}} Z^m\right\|_{\mathrm{F}}^{2} }{\left\|Z^{m\mathrm{T}} Z^m\right\|_{\mathrm{F}}\left\|Z^{k\mathrm{T}} Z^k\right\|_{\mathrm{F}}},
\end{equation}
where $d$ is the dimension of the latent representation, $Z$ is a $n \times d$ matrix of $d$-dimensional representation for $n$ samplesm, $||\cdot||_F$ is the Frobenius norm.

Our results are only evaluated using CKA since we find it~\cite{fedorov2021self} to be the most robust to noise, which reinforces findings in previous literature~\cite{CKA,CKAstructure} that suggest the same.

\subsubsection{Saliency explanation of the representation in brain space}

To explain the representations in brain space, we adapt the
integrated gradients algorithm~\cite{sundararajan2017axiomatic}.
We want to understand the representations rather than the saliency of a specific label. Hence we propose a simple adaptation.
Instead of using a target variable, we compute gradients with respect to each dimension of the representation.
This is done by setting the specific dimension in the vector to 1 and all other dimensions to 0.

\section{Experiments and results}

\begin{figure}[t]
    \centering
    \includegraphics[width=\linewidth]{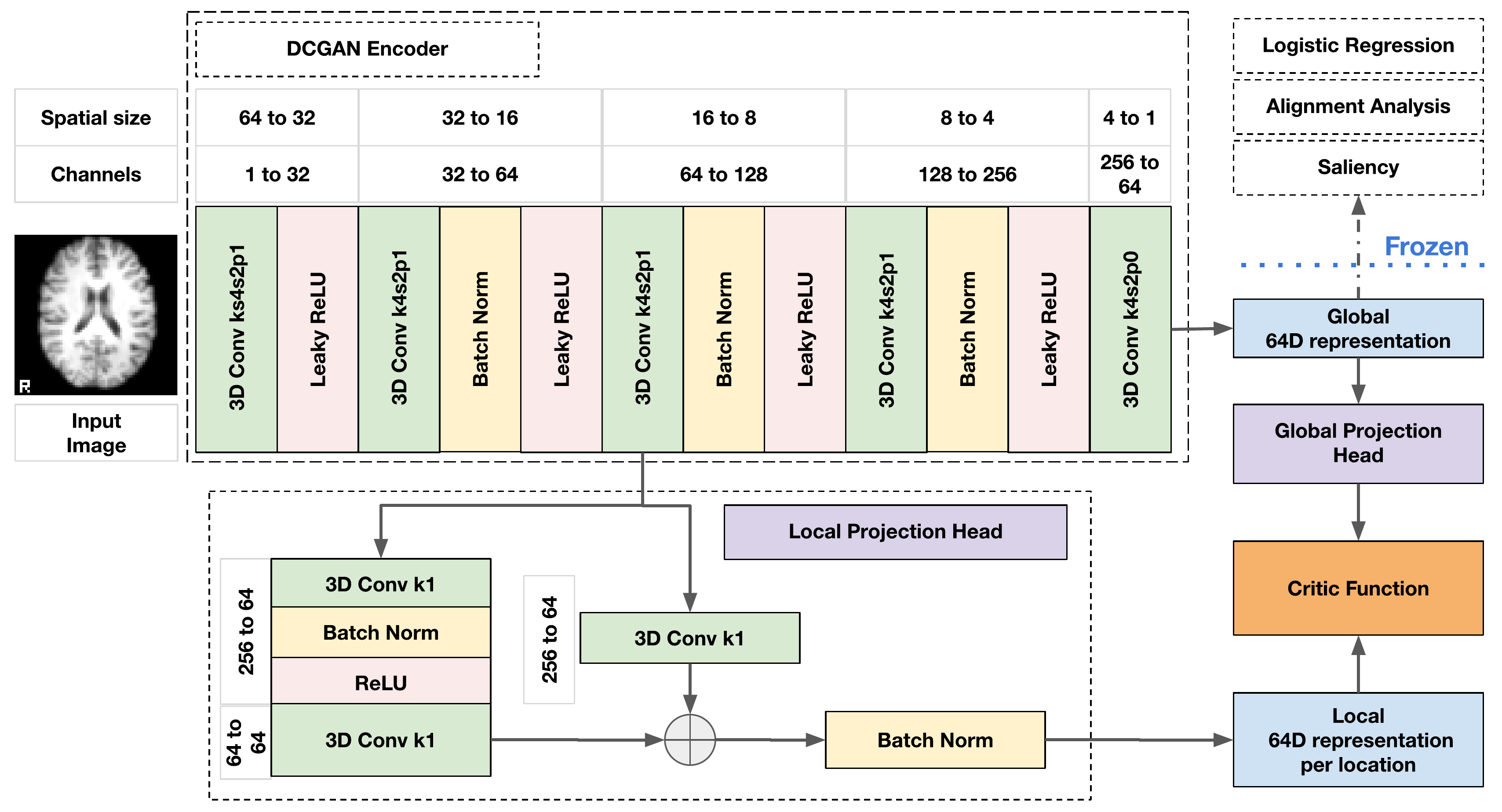}
    \caption{
        The figures show the learning framework for the CR-based objective with the T1 image.
        It includes an encoder with DCGAN~\cite{dcgan} architecture, local and global projection heads, and the computation of the critic function.
        The evaluation of the representation is performed with a frozen pre-trained encoder.
        First, with logistic regression, we evaluate the downstream performance.
        Secondly, with alignment analysis, we explore the multimodal properties of the representation.
        Finally, we interpret the representation in the brain space with saliency gradients.
    }
    \label{fig:framework}
\end{figure}

In this section, we present our findings on two classification tasks with the OASIS-3 dataset, investigating the relative performance of self-supervised and fully supervised approaches.
In addition, to understand the inductive biases of different multimodal objectives, we calculate CKA to measure joint information captured in representations between modalities.
Secondly, we compare group differences between the supervised and best self-supervised model from the first stage.
Lastly, we explore multimodal links between T1 and fALLF.

The overall scheme of our experimental setup is shown in Figure~\ref{fig:framework} which consists of initial pretraining and further evaluation on the classification task, alignment of the representation, and analysis of representation through saliency in brain space.

\subsection{Dataset}

Here we validate our method on OASIS-3~\cite{OASIS3}, which is a multimodal neuroimaging dataset including multiple Alzheimer's disease phenotypes.

Each subject in this dataset is represented by a T1 volume, and a fractional amplitude of low-frequency fluctuation (fALFF)~\cite{zou2008improved} volume, which is generated from T1w and resting-state fMRI (rs-fMRI) images.
The purpose of the T1 volume is to account for the anatomy of the brain, and the fLAFF volume captures the resting-state dynamics. Both T1 volumes and fALFF volumes have previously been shown to be informative for studying not only Alzheimer's disease~\cite{he2007regional} but also other cases (e.g., chronic smokers~\cite{wang2017altered}).

The T1w images were brain-masked with BET in FSL~\cite{fsl} (v 6.0.20), linearly transformed to MNI space, and subsampled to 3mm after preprocessing.
15 T1w images were discarded because these images did not pass the initial visual quality check.
The rs-fMRI was registered using MCFLIRT in FSL~\cite{fsl} (v 6.0.20) to the first image.
The specific parameters for MCFLIRT are: a 3-stage search level (8mm, 4mm, 4mm); 20mm field-of-view, 256 histogram bins (for matching); 6 degrees-of-freedom (DOF) for transformation; a scaling factor of 6mm; the normalized correlation values across the volumes as the cost function (smoothed to 1mm), and interpolation was computed using splines.
The fALFF maps we computed using REST~\cite{rest} are within the 0.01 to 0.1 Hz power band.
The final volume size for both modalities is $64\times64\times64$.
Although it is not required to register the volumes in MNI space, we perform this registration to simplify the analysis and interpretability of our method. Otherwise, we tried to minimize the preprocessing of the data to retain as much information in the original data that the neural network can then learn as possible.
In addition, the subsampling to $3$mm has been done to reduce computational needs, while applications on $1$mm we considered as future work.

Non-Hispanic Caucasian subjects are the largest cohort ($84\%$) in the dataset.
Thus, we selected $826$ ($70\%$ HC, $15\%$ AD, $15\%$ unlabeled) non-Hispanic Caucasian subjects as the main set for the pretraining.
OASIS-3~\cite{OASIS3} contains a large number of subjects that are neither classified as AD nor can readily be called controls. These subjects belong to one of 21 diagnostic categories, including some forms of cognitive impairment, frontotemporal dementia (FTD), Diffuse Lewy body disease (DBLD), and vascular dementia from preclinical cohort and followed longitudinal progression.
We have combined all such subjects into a separate third class.

After matching up the scans of each modality that are closest in date out of all available scans for a subject, the final dataset contains $4021$ pairs.
The $4021$ pairs are split into $5$ stratified folds ($580-582$ subjects ($2828-2944$ pairs), $144-146$ ($653-769$)), and hold-out --- $100$ ($424$).
The number of pairs is greater than the number of subjects because some subjects have multiple scans.
Thus, we utilized more pairs during pretraining but used only one pair of images for each subject in the final evaluation.
For the 2-way classification, we do not use unlabeled data, while for 3-way, we use unlabeled data as a "noisy" phenotypic third-class.

Before feeding the images into the neural network, the intensities of the T1 and fALFF were normalized using min-max rescaling to the unit interval ([0, 1]).
We augment the dataset with random crops of size 64 after reflective padding with size eight from all sides during pretraining.
The decision for the preprocessing and augmentation was based on evaluations of the supervised baseline.
We also considered histogram standardization, z-normalization, random flips, and balanced data sampler~\cite{balanced_data_sampler}. However, the results were not substantially different.
Thus to reduce the computational cost, we use on simple min-max rescaling and random crops.

\subsection{Learning and evaluating representation}

\subsubsection{Pretraining}

To train our model that is schematically shown in Figure~\ref{fig:framework}, we have to choose an architecture for an encoder and global and local projection heads.
The local projection head is needed to project \emph{local} representation to a $64$-dimensional channel space to ensure that the critic scores between the \emph{global} and \emph{local} representations will be computed in the same space.
The global projection head is needed due optimization process.
As it has been shown by authors of SimCLR~\cite{simclr} the last projection to the representation can develop a lower rank condition which is beneficial to the optimization of the objective but can be destructive to the representation.

For our encoder, we choose the architecture from the \emph{deep convolutional generative adversarial networks} (DCGAN)~\cite{dcgan}.
This architecture provides a simple, fully convolutional structure and has a specialized decoder which is important for the performance of generative approaches.
We used volumetric convolutional layers for the experiments with neuroimaging OASIS-3 dataset.
Most of the hyperparameters we left as in the original work~\cite{dcgan}.
We swapped the last tanh activation functions in the decoder with a sigmoid because the intensities of the input images are scaled into the unit interval.
The last layer projects activations from the previous layer to the final $64$-dimensional representation vector that we named \emph{global}.
All convolutional layers are initialized with Xavier uniform~\cite{xavier_uniform} and gain related to the activation function~\cite{paszke2019pytorch}.
Each modality has its encoder with DCGAN architecture.

For a \emph{local projection head} $\ell$ we choose architecture that similar to AMDIM~\cite{amdim}.
The projection head represents one ResNet block from a third $l=3$ layer of DCGAN architecture with feature size $128 \times 8 \times 8 \times 8$.
One direction in the block consists of $2$ convolutional layers (kernel size $1$, number of output and hidden channels $64$, Xavier uniform initialization~\cite{xavier_uniform}).
The second direction consists of one convolutional layer (kernel size $1$, number of output channels $64$, initialization as identity).
The projection heads are individual for each modality, and we added it only if the model has \emph{CC} objective.

For a \emph{global projection head} $g$ we follow SimCLR~\cite{simclr}.
We perform a hyperparameter search for the number of hidden layers in the projection head for each model that can use the projection head (except \emph{Supervised}, \emph{AE}, \emph{CC}).
We have considered cases: without a projection head, with a linear projection head, and a projection head with 1-, 2-, or 3- hidden layers.
The number of output dimensions in the projection layers equals $64$.

In addition, following AMDIM~\cite{amdim} we add regularization to InfoNCE objective by penalizing the squared scores computed by the critic function as $\lambda f(x,y)^2$ with $\lambda = 4\mathrm{e}{-2}$, and cliping the scores by $c\tanh(\frac{s}{c})$ with $c=20$.

We perform the training of the models on OASIS-3 dataset with RAdam~\cite{liu2019variance} optimizer with learning rate ($\text{lr=}4\mathrm{e}{-4}$).
The pretraining step in our framework has been performed for $500$ epochs.
For each trained model, we saved $10$ checkpoints based on the best validation loss.

\subsubsection{Evaluation}

After the pretraining step, to evaluate the discriminative performance of representations learned by the various objectives in our taxonomy, we perform two classification tasks.
The first task is a binary classification of Alzheimer's Disease (AD) vs. Healthy Cohort (HC).
The second task is a ternary classification with an additional phenotypical class.
The first task is easier than the second one because the latter has an added class.

The logistic regression is used to evaluate the discriminative performance of the learned representation of the data.
The logistic regression is trained on \emph{global} representation $z$ after extracting it with a pre-trained encoder.
We use logistic regression (from scikit-learn~\cite{scikit-learn}) to perform classification tasks.
The hyperparameters of the logistic regression were optimized with Optuna~\cite{optuna_2019} for $500$ iterations.
The selections of the hyperparameters are performed based on the validation dataset.
The search space for hyperparameters is defined as follows: inverse regularization strength $C$ is sampled log-uniformly with interval $[1\mathrm{e}{-6}, 1\mathrm{e}{+3}]$, for the elastic net penalty, the mixing parameter is uniformly sampled from unit interval $[0, 1]$.
The logistic regression is trained using \emph{SAGA} solver~\cite{defazio2014saga}.
We use a ROC AUC and one-vs-one (OVO) ROC AUC Macro~\cite{hand2001simple} as a scoring function for a hyperparameter search for binary and ternary classification, respectively.
The OVO strategy for ROC AUC metrics in multiclass classification is computed as the average AUC of all possible pairwise combinations of classes.
In addition, it is insensitive to class imbalance for macro averaging~\cite{scikit-learn}.
Classification is performed separately for each modality by training logistic regression on the representations extracted from that modality using the corresponding convolutional encoder.

After running logistic regression for each of the ten checkpoints, we select the
checkpoint with the maximum ROC AUC metric for a binary case and OVO ROC AUC Macro metric for ternary on the validation set.
After choosing the checkpoint, we choose a $global$ projection head based on cross-validation for some models that need the projection head during pre-training.
Since multimodal models are paired, we select the checkpoint based on average performance on both modalities.
While for unimodal models, we pair checkpoints based on the number of epochs, so the model trained longer together are paired together.

Lastly, the alignment score CKA is computed to measure the joint information content of the representation between modalities as a measure of the inductive bias of the training objective.
The alignment score CKA is also computed on \emph{global} representation $z$.

\subsubsection{Results}

\begin{figure}[ht!]
    \centering
  \includegraphics[width=0.9\linewidth]{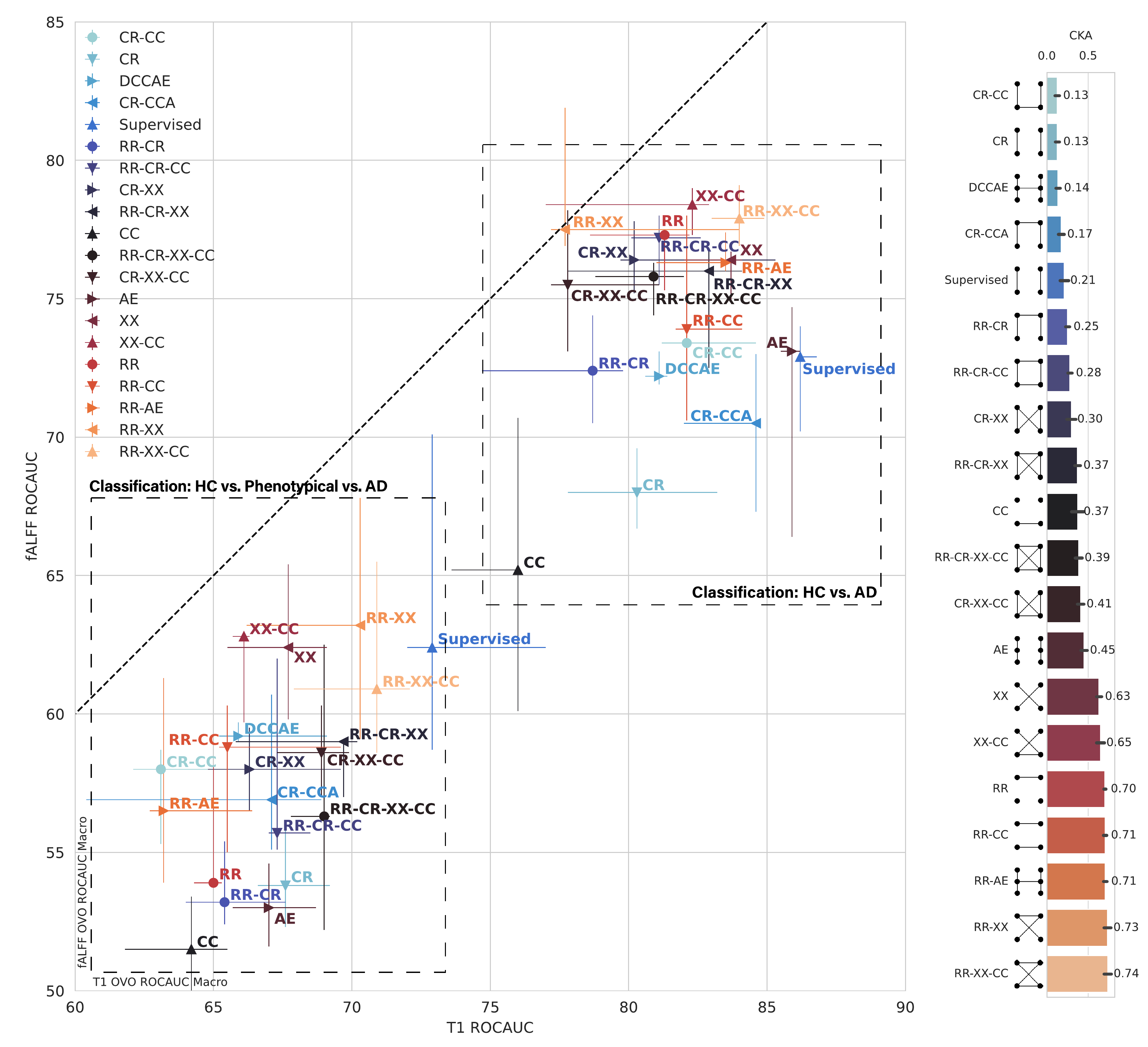}
  \caption{The ROC AUC performance of logistic regression in binary classification (top right corner) and ternary classification (bottom left corner) tasks. Markers correspond to the median of the ROC AUC, and error bars correspond to the IQR. The X- and Y-axis correspond to the ROC AUC on T1 and fALFF modalities. The ROC AUC was measured as a one-versus-one (OVO) macro metric for ternary classification. Two classification tasks are shown on the same plot to visualize the generalizability of the learned representation in tasks with different difficulties. The dashed line represents a diagonal of the balanced performance between T1 and fALFF. The CKA metric shows an alignment of the representation between modalities as a measure of joint information. Lower CKA values mean less joint information between representations, and higher CKA values --- more joint information. Most of the self-supervised models can outperform \emph{Supervised} on both classification tasks. The objectives that include multi-scale local-global relationship \emph{XX} perform robustly on both tasks and retain their ranking relative to the other models. It shows \emph{XX} that is an important building block for multimodal data. The higher content of joint information measured via CKA seems to help models on binary classification but less on harder ternary classification. Thus it does not seem to explain the predictive performance while the multi-scale local-global relationship \emph{XX} of the model can.}
    \label{fig:2way_3way}
\end{figure}

The classification results on a hold-out test dataset are shown for both tasks in Figure~\ref{fig:2way_3way}.
The performance is reported with a median and interquartile range (IQR) of the ROC AUC and one-versus-one (OVO) ROC AUC Macro (average)~\cite{hand2001simple} metrics for binary and ternary classification tasks, respectively.
Additionally, ,we report CKA as the measure of joint information between representations of different modalities for each model.

Overall, the \emph{Supervised} model outperforms self-supervised models on T1 for both tasks: 86.2 (86.1-86.8) for 2-way classification and 72.9 (72.0-77.0) for 3-way classification.
However, the performance gap on T1 is small as the self-supervised multimodal model \emph{RR-XX-CC} achieves 84.0 (83.0-84.9) for binary and 70.9 (67.9-72.1) for ternary classifications.

Most self-supervised models can achieve better classification
performance on "noisier" fALFF than \emph{Supervised} model in the 2-way classification task.
For 3-way classification, the gap is reduced while the \emph{RR-XX} model achieves 63.2 (59.1-67.8) versus 62.4 (58.7 70.1) for \emph{Supervised}.
This supports the benefits of multimodal learning, which could be seen as a regularization effect.

The unimodal autoencoder (\emph{AE}) model can perform well on the simple binary classification task. However, the performance significantly drops on the harder ternary classification task.
Unimodal models such as \emph{CR} and \emph{AE} are outperformed by most of the multimodal models.
Evidently, multimodal extension of \emph{AE} with \emph{CCA} as \emph{DCCAE} can improve performance.
However, \emph{DCCAE} can be outperformed by most of self-supervised decoder-free models on 2-way classification task and \emph{XX-CC}, \emph{RR-XX}, \emph{RR-XX-CC}, \emph{XX} on 3-way classification task.
Therefore we could achieve more robust performance with the proposed models while reducing the computational cost of the decoder for each modality.

Overall, the proposed self-supervised models such as \emph{XX},
\emph{XX-CC}, \emph{RR-XX} and \emph{RR-XX-CC} perform robustly on
both tasks and retain their ranking relative to the other models.
Additionally, judging by the higher CKA alignment measure (0.63-0.73), these models capture joint information between modalities.
While there are other models---\emph{RR-AE}, \emph{RR-CC} and \emph{RR}---that achieve higher CKA alignment yet are not as robust.
We hypothesize that the joint information alone is not the answer to
the problem, but the architecture of the model is important.
Note that the \emph{XX}, \emph{XX-CC}, \emph{RR-XX} and \emph{RR-XX-CC} models capture local-global relationship between modalities.
While the \emph{RR-AE}, \emph{RR-CC} and \emph{RR} models only capture joint information on global-global or local-local representation level.
Thus, given the empirical evidence in Figure~\ref{fig:2way_3way}, the local-global relationship \emph{XX} is an essential building block for multimodal data because it allows us to capture complex multi-scale relationships between modalities.

\subsection{Interpretability}

\subsubsection{Explaining group differences between HC and AD}
In this subsection, we explain the performance of the models by analyzing the saliency maps.
As a point of interest and comparison, we select \emph{Supervised} and \emph{RR-XX} models.
The \emph{Supervised}model performs the best with T1 input volumes and utilizes target labels. Thus it is a solid baseline to analyze group differences.
The \emph{RR-XX} model performs the best in the ternary classification task and precisely does well for fALFF input volumes.
We use these two models to generate saliency maps and interpret what the models have learned.

For each selected model, we compute integrated gradients~\cite{sundararajan2017axiomatic} along each dimension in the 64-dimensional representation and discard the negative gradients.
After computing saliency gradients for each dimension of the latent representation, we apply brain masking, rescale gradient values to a unit interval and smooth them with a Gaussian filter ($\sigma=1.5$).
Then we perform a voxel-wise two-sided test with Mann-Whitney U-Test and compute Rank Bisseral Correlation (RBC) as an effect size.
After selecting the voxels with a $p \in [0., 0.025] \cup [0.975, 1.]$, using \emph{3dClusterize} from AFNI~\cite{cox1996afni}, we find clusters with at least 200 voxels.
Then we apply \emph{whereami} from AFNI~\cite{cox1996afni} to match those clusters with the ROIs defined in the template that is used in the Neuromark pipeline~\cite{du2020neuromark}.
We call this template the Neuromark atlas in the rest of the text.
To create the Neuromark atlas, the spatial ICA components of the Neuromark template have been combined to create an atlas by simple overlapping.
Then atlas has been added to AFNI~\cite{cox1996afni} environment.

\begin{figure}[ht!]
    \centering
    \includegraphics[width=\linewidth]{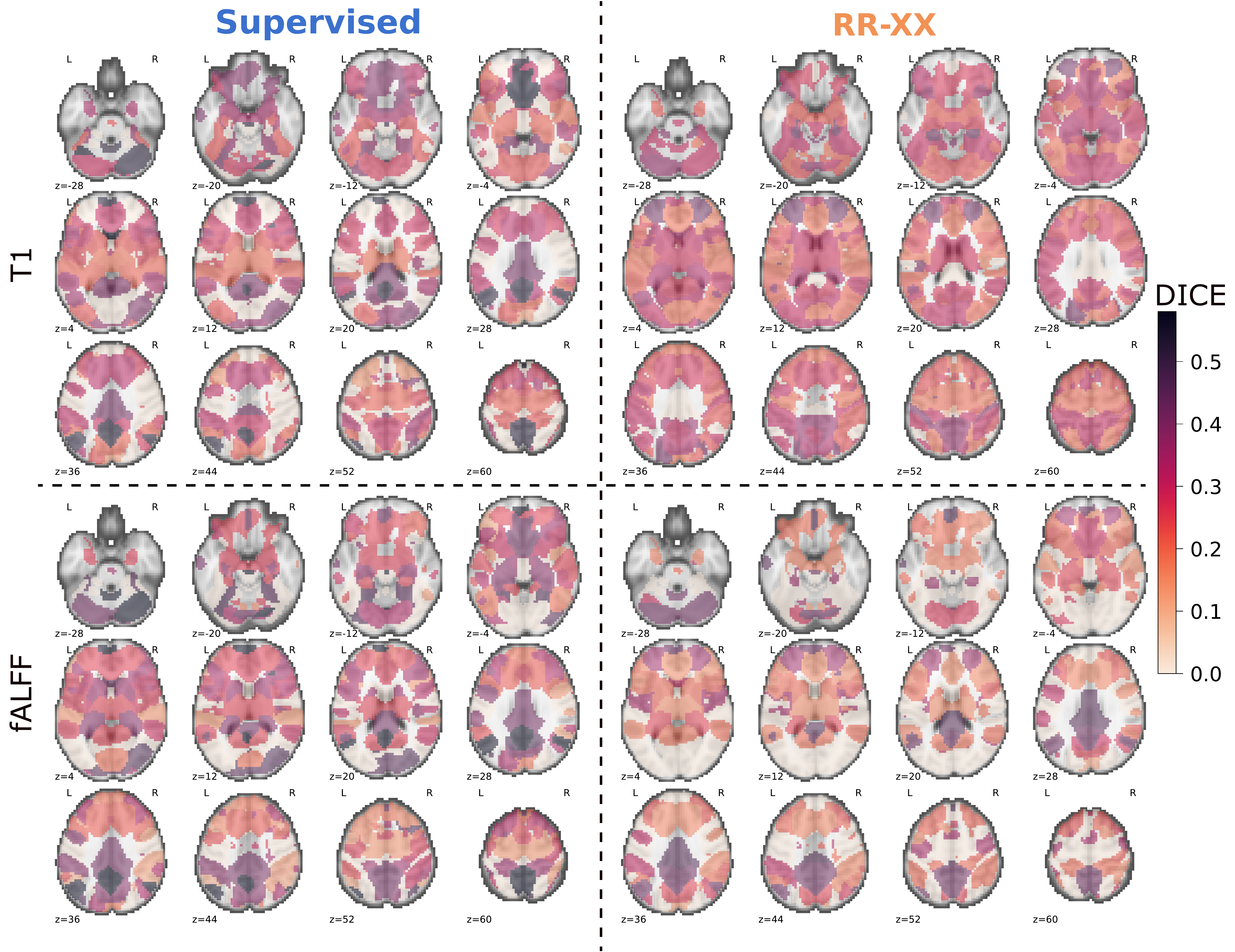}
    \caption{The figure shows the maximum DICE overlap of the saliency clusters with regions in the Neuromark atlas. The first row shows the DICE maps for sMRI data, and the second row shows the maps found for fALFF. The first columns show maps for the \emph{Supervised} model, and the second column shows the maps for the \emph{RR-XX} model. \emph{Supervised} shows a sparse choice of discriminative regions on T1, overall higher DICE performance, and stronger contrast. \emph{RR-XX} shows a sparse choice of discriminative regions on fALFF, overall lower DICE, and lower contrast.}
    \label{fig:DICE}
\end{figure}

We select only the top brain ROI for each cluster based on the overlap and include only the ROIs that have been found in at least two folds.
The overlap in the clustered saliencies with ROIs is measured using DICE.
The final results are summarized for all dimensions in Figure~\ref{fig:DICE} where we report the maximum DICE overlap for both models and both modalities.

The results in Figure~\ref{fig:DICE} suggest that higher discriminative performance is related to a sparse choice of the ROIs.
Specifically, the \emph{Supervised} model seems to be sparser than the \emph{RR-XX} model on T1 volumes because its binary ROC AUC performance is $86.2$, compared to $77.7$.
The \emph{RR-XX}, however, seems sparser than the \emph{Supervised} for fALFF data because its 2-way ROC AUC performance is $77.5$ compared to $72.9$.

Another results in Figure~\ref{fig:DICE} that the \emph{Supervised} model has higher DICE and stronger contrast compared to the \emph{RR-XX}.
It suggests stronger localization of the saliency maps for \emph{Supervised} model that can be explained by the use of the labels to learn representation.
As a self-supervised model learns without labels, a lower contrast can support the idea that self-supervised learning learns less task-specific general representation.

In addition, in Figure~\ref{fig:DICE}, the models tend to capture information from the frontal lobe regions and less of the posterior part of the brain in their representations on fALFF.

\subsubsection{Detailed analysis of group differences with best self-supervised and supervised models}

\begin{figure}[ht!]
    \centering
    \includegraphics[width=0.95\linewidth]{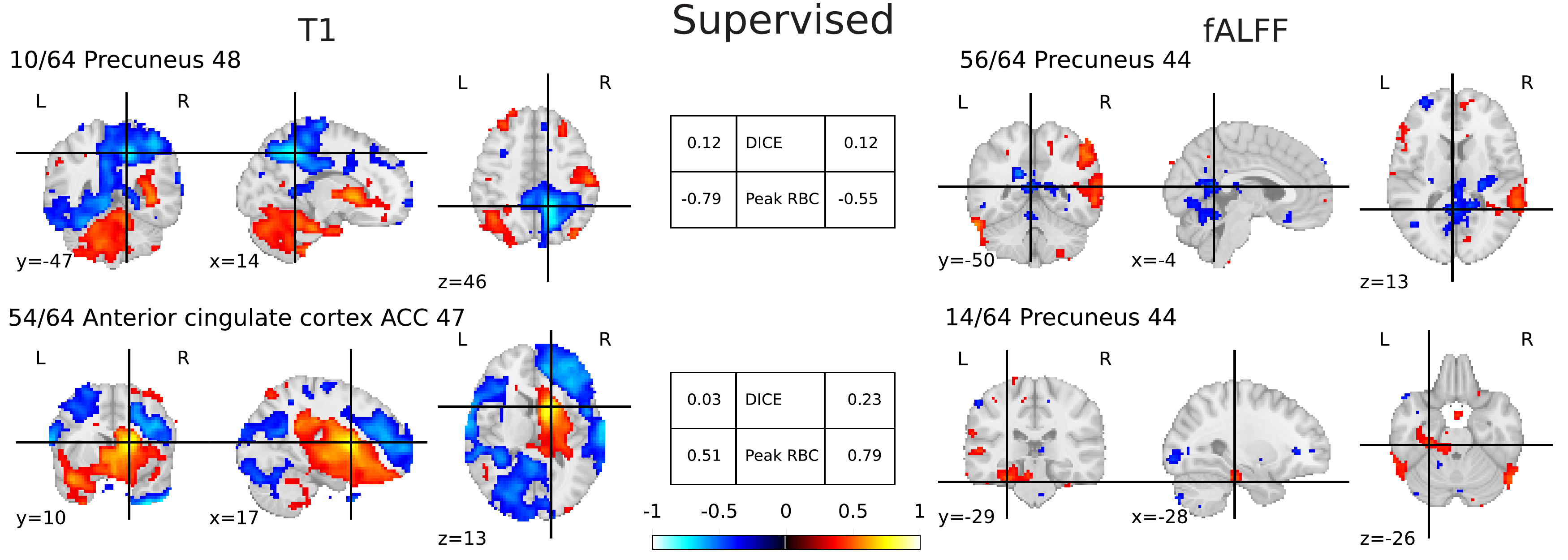}
    \caption{The figures show the regions of interest of the Neuromark atlas that correspond to clusters (>200 voxels) in voxel-wise RBC maps with the highest DICE overlap for \emph{Supervised} model. We show a peak RBC value of the cluster. The maps are chosen for dimensions in the representation that corresponds to the highest positive and lowest negative betas in the trained logistic regression. We show the regions that are found for T1 in the left column and the right column for fALFF. \emph{Supervised} finds precuneus and anterior cingulate cortex.}
    \label{fig:Supervised-group-differences}
\end{figure}

\begin{figure}[ht!]
    \centering
    \includegraphics[width=\linewidth]{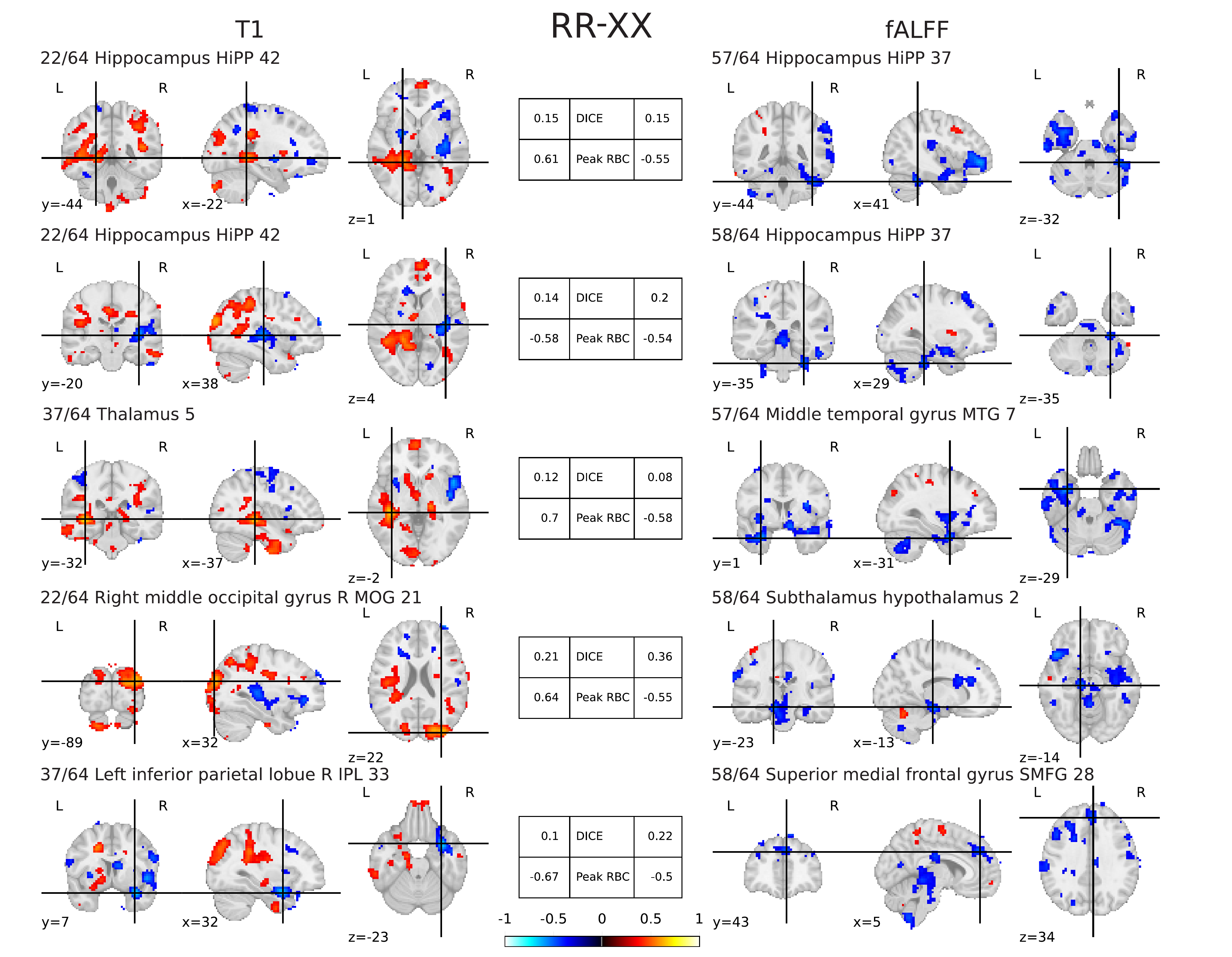}
    \caption{The figures show the regions of interest for the Neuromark atlas that corresponds to clusters ($>200$ voxels) in voxel-wise RBC maps with the highest DICE overlap for \emph{RR-XX} model. We show a peak RBC value of the cluster. The maps are chosen for dimensions in the representation that corresponds to the highest positive and lowest negative betas in the trained logistic regression. We show the regions that are found for T1 in the left column and the right column for fALFF. \emph{RR-XX} finds many discriminative regions that are supported by the literature, such as the hippocampus, thalamus, parietal lobule, occipital gyrus, middle temporal gyrus, superior medial frontal gyrus, subthalamus hypothalamus.}
    \label{fig:RRXX-group-differences}
\end{figure}

In this analysis, we compare saliency maps for group differences for \emph{Supervised} and \emph{RR-XX} models specifically for most discriminative dimensions in the representation vector.
The discriminative dimensions are the dimensions in $z$ with the highest and lowest beta scores in the trained logistic regression.
After selecting the dimensions, we compute saliencies and perform a voxel-wise test as in the previous subsection to get RBC maps with significant voxels.

In Figure~\ref{fig:Supervised-group-differences} we show RBC maps for \emph{Supervised}model, and in Figure~\ref{fig:RRXX-group-differences} we show RBC maps for \emph{RR-XX} for only the first fold on the hold-out test set.

The \emph{Supervised} model has bigger clusters on T1, while the self-supervised model \emph{RR-XX} has more local and smaller clusters.
Given that \emph{Supervised} has better performance on T1 than \emph{RR-XX}, these RBC maps might explain the performance gap in 2-way classification.
However, given the reduced gap in 3-way classification, it might also indicate that the \emph{Supervised} model might overfit the task and uses more regions than is needed.

As we can see in Figure~\ref{fig:RRXX-group-differences}, the self-supervised model \emph{RR-XX} able to pick on T1 the discriminative regions that supported by the literature: such as hippocampus~\cite{yang2022human}, thalamus~\cite{elvsaashagen2021genetic}, parietal lobule~\cite{greene2010subregions}, occipital gurys~\cite{yang2019study}.
The regions that are found for \emph{Supervised}are also supported by the literature: such as precuneus~\cite{guennewig2021defining} and anterior cingulate cortex~\cite{yu2021human}.

On fALFF, \emph{Supervised} and \emph{RR-XX} models share similar behavior, including more local and smaller clusters.
The \emph{supervised} model shows precuneus, consistent with prior work focused on fALFF~\cite{wang2021comparative}.
The \emph{RR-XX} model shows hippocampus~\cite{10.3389/fnagi.2018.00037,zhang2021regional}, middle temporal gyrus~\cite{hu2022brain}, subthalamus hypothalamus~\cite{chan2021induction} and superior medial frontal gyrus~\cite{cheung2021diagnostic}.

\begin{figure}[ht!]
    \centering
    \includegraphics[width=\linewidth]{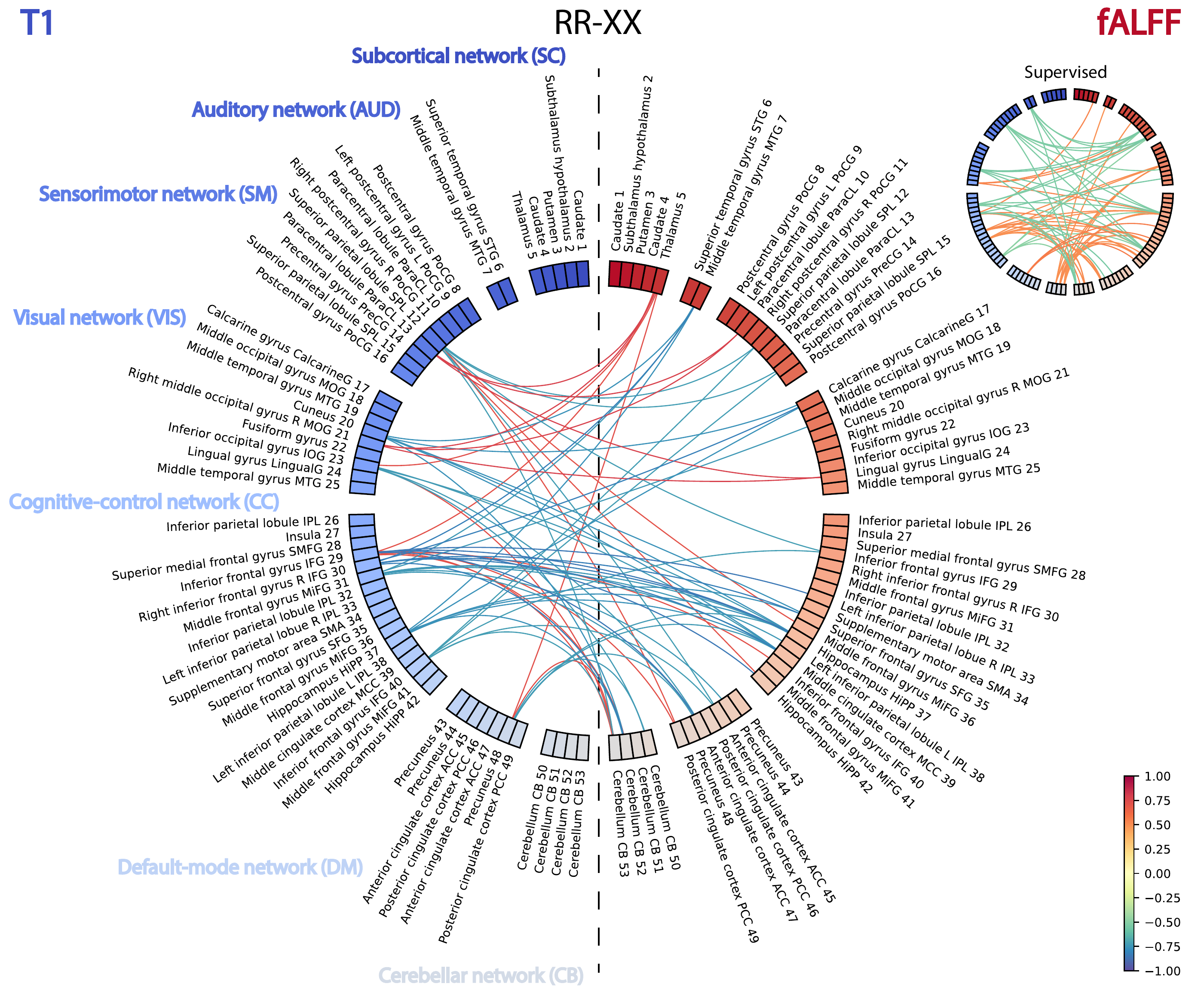}
    \caption{The Figure shows multimodal links between T1 and fALFF of ROIs in the Neuromark atlas for \emph{RR-XX} model while for \emph{Supervised} in the left top corner. The ROIs for T1 are shown on the left side with shades of blue, and the ROIs for fALFF are shown on the right side with shades of red. The edge weights are defined by the correlation between dimensions in the representation vector between T1 and fALFF and colored according to the spectral color bar. \emph{RR} finds multiple multimodal links between regions supported by the literature, such as thalamus-precuneus, precuneus-hippocampus, and precuneus-middle cingulate cortex.}
    \label{fig:RRXX-alignment}
\end{figure}

\subsubsection{Exploring multimodal links}

This section explores multimodal links between the T1 and fALFF modalities.
To perform this analysis, we compute an asymmetric correlation matrix between all pairs of dimensions in $64$-dimensional \emph{global} representation of the T1 and fALFF.
Then we select one ROI in the Neuromark atlas and find a dimension in the representation vector with a cluster with the highest DICE overlap with this ROI in RBC maps.
After finding this dimension, we find a second dimension from another modality with the highest positive and negative correlation from the correlation matrix.
Then we connect the first ROI with each ROI captured by a second dimension with an edge with correlation values as a weight.
We repeat the same procedure for each of the 53 ROIs in the Neuromark cluster and each modality.

The final summarization of the multimodal relationships is shown in Figure~\ref{fig:RRXX-alignment}.
Note, we show the top 64 edges with maximum by absolute values weights, and, specifically, we only focus on the self-supervised multimodal model \emph{RR-XX}.
However, we also show the same diagram for the \emph{Supervised} model in a restricted way.
Because the correlation of the \emph{Supervised} is much lower, and the model is learning representation unimodally, thus the relationships are more likely to be spurious.
In the figure, we show ROIs for T1 on the left side with blue hues and fALFF on the right side with red hues.
Additionally, we group ROIs by functional networks defined in the Neuromark atlas.

One positively correlated link (Pearson's $r(86)=.724, p<0.001$) has been found by \emph{RR-XX} is thalamus 5 (fALFF) - precuneus 48 (T1) that has been associated with changes in consciousness~\cite{cunningham2017structural}.
Another negatively correlated link (Pearson's $r(86)=-.511, p<0.001$) with precuneus 48 (T1) and hippocampus 37 (fALFF) has been associated with Alzheimer's disease~\cite{kim2013hippocampus, ryu2010measurement}.
The negatively correlated link (Pearson's $r(86)=-.539, p<0.001$) between precuneus 48 (T1) and middle cingulate cortex 39 (fALFF) could be related with findings in these works~\cite{rami2012distinct, bailly2015precuneus}.

Overall, the self-supervised model \emph{RR-XX} can learn meaningful multimodal relationships that clinicians can further explore.

\subsection{Hardware, reproducibility, and code}

The experiments were performed using an NVIDIA V100.
The code is implemented mainly using PyTorch~\citep{paszke2019pytorch} and Catalyst~\citep{catalyst} frameworks.
The code is available at \href{https://github.com/Entodi/fusion}{github.com/Entodi/fusion} for reproducibility and further exploration by the scientific community.

\section{Discussion}

\subsection{Multi-scale coordinated self-supervised models}

The proposed self-supervised multimodal multi-scale coordinated models can capture useful representation and multimodal relationships in the data.
Compared to existing unimodal (\emph{CR} and \emph{AE}) and multimodal (\emph{DCCAE}) counterparts, these models achieve higher discriminative performance in ROC AUC on downstream tasks. While some of them can capture higher joint information content between modalities as measured by CKA.
Furthermore, these models can produce representations that, compared to the \emph{Supervised} model, show competitive performance on T1 and outperform fALLF.

We show strong empirical evidence that the \emph{XX} is the most important relationship to encourage when high discriminative performance is the goal.
This results is evidence of the importance and existence of multi-scale local-to-global multimodal relationships in the functional and structural neuroimaging data, which other relationships can not capture.

However, not all multimodal variants from the taxonomy of Figure~\ref{fig:variants} result in robust and useful representations.
Specifically, our experiments show that the \emph{CC} relationship should not be used separately from other objectives, as the \emph{CC} will optimize only the layers below the chosen one because the last layer will behave as a random projection.
We show it only for a complete picture of achievable classification performance with all objectives in the taxonomy.

The CCA-based objectives did not show the good results as would be expected based on the current literature.
However, our taxonomy revealed that the \emph{DCCA} is related to \emph{SimCLR}~\cite{simclr} which led us to develop the \emph{RR} model.
While CCA maximizes correlations, \emph{SimCLR} maximized cosine similarity between representations of different modalities.
However, the \emph{SimCLR} objective has one more important difference: it performs an additional discrimination step on cosine similarity scores.
Thus it does two things: maximizes the similarity between modalities and simultaneously performs additional discrimination of pairs based on similarity.
This task is more challenging because it needs to capture richer information to classify pairs of representations from different modalities based on similarity.
In addition, the CCA-objective is prone to numerical instability thanks to its implementation in DCCAE~\cite{DCCAE}.
\emph{RR} does not have such issues.
We recommend using ``softer'' optimization based on mutual information estimators with deep neural networks and not the ``exact'' solutions based on linear algebra in DCCAE~\cite{DCCAE}.

While \emph{AE} imposes an additional computational complexity due to the decoder, it has not shown benefits to the discriminative performance of the model.
Specifically, \emph{AE} struggles to deal with ternary classification tasks.
Multimodal models from our proposed taxonomy have a reduced computational burden lacking a volumetric decoder.
These findings concur with the poor performance of autoencoders on datasets with natural images~\cite{DIM}.
We hypothesize that autoencoders, to achieve greater performance, may require encoders and decoders of high capacity.
However, this will considerably increase the difficulty of training large volumetric models.

\subsection{Future Work}

The models we have constructed in this work do not disentangle representation into joint and unique modality-specific representations.
The analysis between CKA and downstream performance shows existence of a joint subspace between modalities, and a specific amount of joint information measured by CKA is important to learn representation valuable for downstream tasks.
Future work could consider models that explicitly represent factors of the joint and unique components.
Some related ideas have been explored in work on natural images when disentangling content and style~\cite{von2021self,lyu2021understanding}, similarly, for neural data with variational autoencoders~\cite{liu2021drop}.

In our analysis, we do not consider the family of multimodal generative variational models~\cite{kingma2013auto}.
Currently, volumetric variational models are computationally expensive, and the field is under active development given many models that have been proposed recently.
Including all possible models with all possible underlying technology was not precisely our goal and would make the already extensive list of models hard to analyze.
Future work may consider variational models under the same taxonomy for a fair comparison and detailed analysis of multimodal fusion applications.

There is more than can be done concerning the explainability of the models.
Currently, a common choice to model neuroimaging data is to use a convolutional neural network (CNN)~\cite{abrol2021deep}.
However, the simple application of CNNs leads to representation, where each dimension captures multiple ROIs.
This effect creates difficulties in analyzing the cross-modal relationship between modalities. The multimodal links between ROIs can only be measured by the correlation between dimensions of a representation in different modalities.
Thus the measured links do not represent the multimodal link between one ROI and another ROI but rather between dimensions.
Future work may consider ROI-based representations.

In addition, as we want to focus on unsupervised models without using group labels, we used HC and AD groups to show group differences or identify ROIs in our analysis.
However, the data may contain phenotypically small groups of patients that are not represented by HC or AD groups.
It will be hard to do group analysis in such a scenario because we do not have the labels.
Thus future work can consider additional clustering of the representation for finding such subgroups that explainability methods can further analyze.

\section{Conclusions}

In this work, we presented a novel multi-scale coordinated framework for representation learning from multimodal neuroimaging data.
We showed that self-supervised approaches can learn meaningful and useful representations which capture regions of interest with group differences without accessing group labels during the pre-training stage.
We developed evaluation methodologies to access the properties of representations learned by
models within the family of models in downstream task analysis,
measurements of joint subspace, and explainability evaluations.

We outperformed previous unsupervised models AE and DCCAE on all classification tasks and modalities.
In addition, our family of models does not require a decoder that saves computational and memory requirements.
In addition, we can outperform the \emph{Supervised} model on fALFF. This result suggests future use of the proposed multimodal objectives for asymmetric fusion as a regularization technique.
Further, our findings suggest the importance of multi-scale local-to-global
multimodal relationships \emph{XX} =
\includegraphics[valign=c,width=0.7\baselineskip]{XX} that
considerably improve the performance and multimodal alignment over previous methods and within the proposed family of models.
This result suggests that there exist multi-scale relationships between local structure and global summary of the inputs in different modalities that previously have been neglected in multimodal representation learning.

The \emph{RR-XX} model, selected based on the best classification performance and higher joint information content via CKA, was able to capture important regions of interest related to Alzheimer's disease such as hippocampus~\cite{yang2022human,10.3389/fnagi.2018.00037,zhang2021regional}, thalamus~\cite{elvsaashagen2021genetic}, parietal lobule~\cite{greene2010subregions}, occipital gurys~\cite{yang2019study} , middle temporal gyrus~\cite{hu2022brain}, subthalamus hypothalamus~\cite{chan2021induction}, and superior medial frontal gyrus~\cite{cheung2021diagnostic}.
Importantly, the \emph{RR-XX} model is able to capture multimodal links between regions that are supported by the literature such as thalamus-precuneus~\cite{cunningham2017structural}, precuneus-hippocampus~\cite{kim2013hippocampus, ryu2010measurement}, and precuneus middle cingulate cortex~\cite{rami2012distinct, bailly2015precuneus}.

The showcased benefits of applying a comprehensive approach, evaluating a taxonomy of methods, and performing extensive qualitative and quantitative evaluation suggest that multimodal representation learning is a field with significant potential in neuroimaging, despite being in a nascent state. Our work lays a foundation for future robust and increasingly more interpretable multimodal models.

\section*{Acknowledgments}
This work was funded by the National Institutes of Health (NIH) grants R01MH118695, RF1AG063153, 2R01EB006841, RF1MH121885, and the National Science Foundation (NSF) grant 2112455.

Data were provided by OASIS-3: Principal Investigators: T. Benzinger, D. Marcus, J. Morris; NIH P50 AG00561, P30 NS09857781, P01 AG026276, P01 AG003991, R01 AG043434, UL1 TR000448, R01 EB009352. AV-45 doses were provided by Avid Radiopharmaceuticals, a wholly owned subsidiary of Eli Lilly.

\typeout{}
\bibliography{references}
\end{document}